\documentclass[conference]{IEEEtran}
\IEEEoverridecommandlockouts

\makeatletter
\def\endthebibliography{%
	\def\@noitemerr{\@latex@warning{Empty `thebibliography' environment}}%
	\endlist
}

\usepackage{cite}
\usepackage{amsmath,amssymb,amsfonts}
\usepackage{algorithmic}
\usepackage{graphicx}
\usepackage{textcomp}
\usepackage{xcolor}
\usepackage{multirow}

\usepackage{hyperref}
\usepackage{cleveref}

\newcommand{\andothers}{\MakeLowercase{\textit{ et al. }}}
\begin{document}

\title{Sign Language Recognition Analysis using Multimodal Data\\
}
\author{\IEEEauthorblockN{Al Amin Hosain,
Panneer Selvam Santhalingam, Parth Pathak, Jana Ko\v{s}eck\'a and Huzefa Rangwala}
\IEEEauthorblockA{\textit{Department of Computer Science} \\
\textit{George Mason University} \\
Fairfax, USA\\
\{ahosain, psanthal, phpathak, kosecka, rangwala\}@gmu.edu}}

\maketitle

\begin{abstract}
Voice-controlled personal and home assistants (such as the Amazon Echo and Apple Siri) are becoming increasingly popular for a variety of applications. However,  the benefits of  these technologies 
		are not readily accessible to Deaf or Hard-of-Hearing (DHH) users. 
		%
		The objective of this study is to
		develop and evaluate a sign recognition system using multiple  modalities that can be used by DHH signers to interact with voice-controlled devices. 
		With the advancement of depth sensors,  skeletal data is used for applications like video analysis and activity recognition. Despite having similarity with the well-studied human activity recognition, the use of 3D skeleton data in sign language recognition is rare. 
		This is because unlike activity recognition, sign language is mostly dependent on hand shape pattern. %
		In this work, we investigate the feasibility of using skeletal and RGB video data
		for sign language recognition using a combination of different deep learning architectures.  
		We validate our results on a large-scale American Sign Language (ASL) dataset of 12 users and 13107 samples across 51 signs. It is named as GMU-ASL51. \footnote{Contact author for the details.} 
		We collected the dataset over 6 months and it will be publicly released in the hope of spurring further machine learning research towards providing improved accessibility for digital assistants. \footnote{IRB protocol 1338304-2: For privacy purposes we can only make the skeletal data and hand patches publicly available.}  
		%
		%
		%

\end{abstract}

\begin{IEEEkeywords}
neural networks, deep learning, modality-fusion, sign language recognition
\end{IEEEkeywords}

\section{Introduction}

According to The National Institute on Deafness, one in thousand infants is born deaf. An additional one to six per thousand are born with hearing loss at different levels~\cite{3072291}. Sign language is commonly used by Deaf and Hard-of-Hearing (DHH) persons to communicate via hand gestures.  An automatic sign language recognizer enables an ASL user to translate the sign language to written text or speech,  allowing them to communicate with people who are not familiar with ASL. There is a tremendous rise in the popularity of personal digital assistants; available on user's personal and wearable devices (Google Now, Amazon Alexa and Apple Siri, etc.) and also in the form of standalone devices (Amazon Echo and Google Home smart speakers). These devices are primarily controlled through voice, and hence, their functionality is not readily available to DHH users. An automatic sign recognizer can also enable the interaction between a DHH user and a digital assistant. 

	Most current systems have capability of ASL recognition with RGB video data \cite{7177428,Ji:2013:CNN:2412386.2412939,Sun:2015:LSV:2753829.2629481}. An ASL sign is performed by a combination of hand gestures, facial expressions and postures of the body. Sequential motion of specific body locations (such as hand-tip, neck and arm) provide informative cues about a sign.
	Using video data, it is difficult to extract different body locations and associated motion sequences from a series of RGB frames. Microsoft Kinect is a 3D camera sensor which can use the depth information of a person to capture 3D coordinates of his/her body location across a video. This sequence of 3D body location is referred by skeletal data~\cite{Zhang:2012:MKS:2225053.2225203}. 
	To the best of our knowledge, there is no publicly available skeletal dataset in literature for ASL recognition. 
	
	With skeletal data, an ASL sign can be seen as a sequence of 3D coordinates or a 3D time series~\cite{7298714}. 
	Recurrent neural networks (RNN) have shown strong performance for sequential modeling~\cite{DBLP:journals/corr/Lipton15}. 
	In this work, we investigate the impact of RGB video data in recognition accuracy when combined with skeletal data. We also propose a combined RNN network with a simple spatial data augmentation technique. 
	In summary, the contributions of this work are:
	\begin{itemize}
		\item We propose an RNN architecture with a novel spatial data augmentation technique.  
		\item We propose an architecture which uses both RGB and skeletal data to improve recognition accuracy.
		\item We introduce and publicly release a multi--modal dataset for ASL called GMU-ASL51.
	\end{itemize}

\section{Literature Review}
Most sign language recognition systems use RGB video data as input. These approaches model sequential dependencies using Hidden Markov Models (HMM). Zafrullah\andothers\cite{copycat_zafrulla} used colored gloves (worn on hands) during data collection and developed an HMM based framework for ASL phrase verification. They also used hand crafted features from Kinect skeletal data and accelerometers worn on hand \cite{Zafrulla:2011:ASL:2070481.2070532}. 
	Huang\andothers\cite{7177428} demonstrated the effectiveness of using Convolutional neural network (CNN) with RGB video data for sign language recognition. Three dimensional CNN have been used to extract spatio-temporal features from video \cite{Ji:2013:CNN:2412386.2412939}. Similar architecture was implemented for Italian gestures \cite{978-3-319-16178-5_40}. Sun\andothers\cite{Sun:2015:LSV:2753829.2629481}
	hypothesized that not all RGB frames in a video are equally important and assigned a binary latent variable to each frame in training videos for indicating the importance  of a  frame within a  latent support vector machine model. Zaki\andothers\cite{ZAKI2011572} proposed two new features with existing hand crafted features and developed the system using HMM based approach. Some researchers have used appearance-based features and divided the approach into sub units of RGB and tracking data, with a HMM model for recognition \cite{Cooper2017}. 

	Compared to RGB methods, skeletal data has received little attention in ASL recognition.
 	However, in a closely similar human action recognition task, a significant amount of work has been done using body joint location related data. Shahroudy\andothers\cite{7780484} published the largest dataset for human activity recognition. They proposed an extension of long short term memory (LSTM) model which leverages group motion of several body joints to recognize human activity from skeletal data. A different adaptation of the LSTM model was proposed by Liu\andothers\cite{8101019} where spatial interaction among joints was considered in addition to the temporal dynamics. Veeriah\andothers\cite{7410817} proposed a LSTM network to capture the salient motion pattern of body joints. This method takes into account the derivative of motion states associated with different body joints. Some have treated the whole body as a hierarchical configuration of different body parts and proposed a hierarchical RNN to recognize human activities \cite{7298714}. Several attention based models were  proposed for human activity analysis \cite{song2016end,8226767}. Some prior works converted skeleton sequences of body joints or RGB videos into an image representation and then applied state-of-the-art image recognition models \cite{DBLP:journals/corr/abs-1711-05941,DBLP:conf/cvpr/KeBASB17}.
	Motivated by the success of skeletal data in human activity recognition, we investigate its suitability for recognizing ASL signs. 
\section{Dataset}
	ASL recognition with skeletal data has received little attention, resulting in a scarcity of public datasets. There exists one dataset for ASL recognition with skeletal data \cite{4563181}. This dataset has 9800 samples from 6 subjects and more than 3300 sign classes. The number of samples per class was small for use in deep learning based models. Adding to this, the samples were collected in controlled settings with uncluttered background. 
	In contrast, GMU-ASL51 has 13107 samples for 51 word level classes from 12 distinct subjects of different height, build and signing (using sign language) experience.
	\begin{figure}[h]
		\begin{center}
			\includegraphics[width=.8\linewidth]{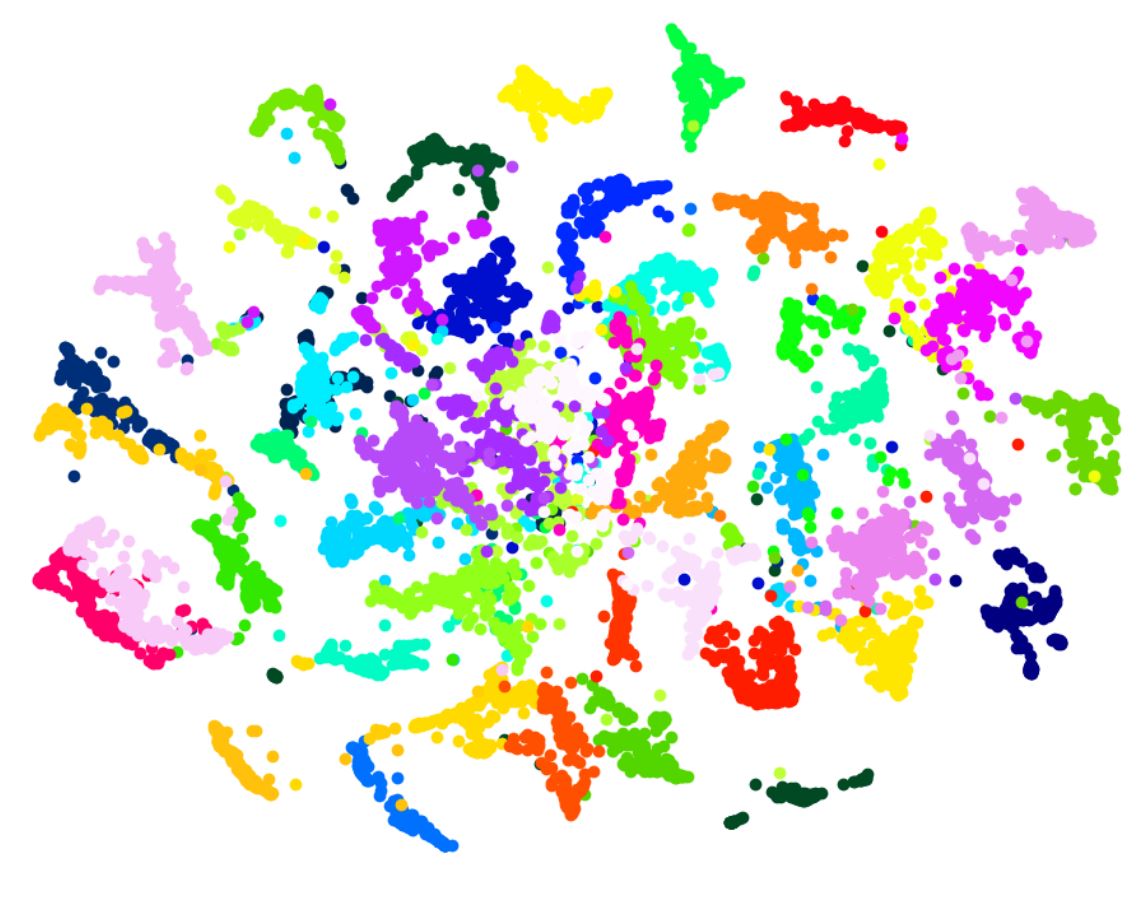}
		\end{center}
		\caption{T-SNE representation of 11824 data samples from 51 different sign classes. Best viewed in color.}
		\label{fig:tsne_rep}
	\end{figure} 
	Figure \ref{fig:tsne_rep} shows the T-SNE representation of a subset of samples from GMU-ASL51. It was performed on output vectors from a trained RNN model for each sign example in the subset. The used model, AI-LSTM, is described in section \ref{axis_ind_rnn}.
	\begin{figure*}[h]
	\begin{center}
		\includegraphics[clip, trim= 0.1cm 0.7cm 0.1cm 0.0cm, width=\linewidth, height=.3\textheight]{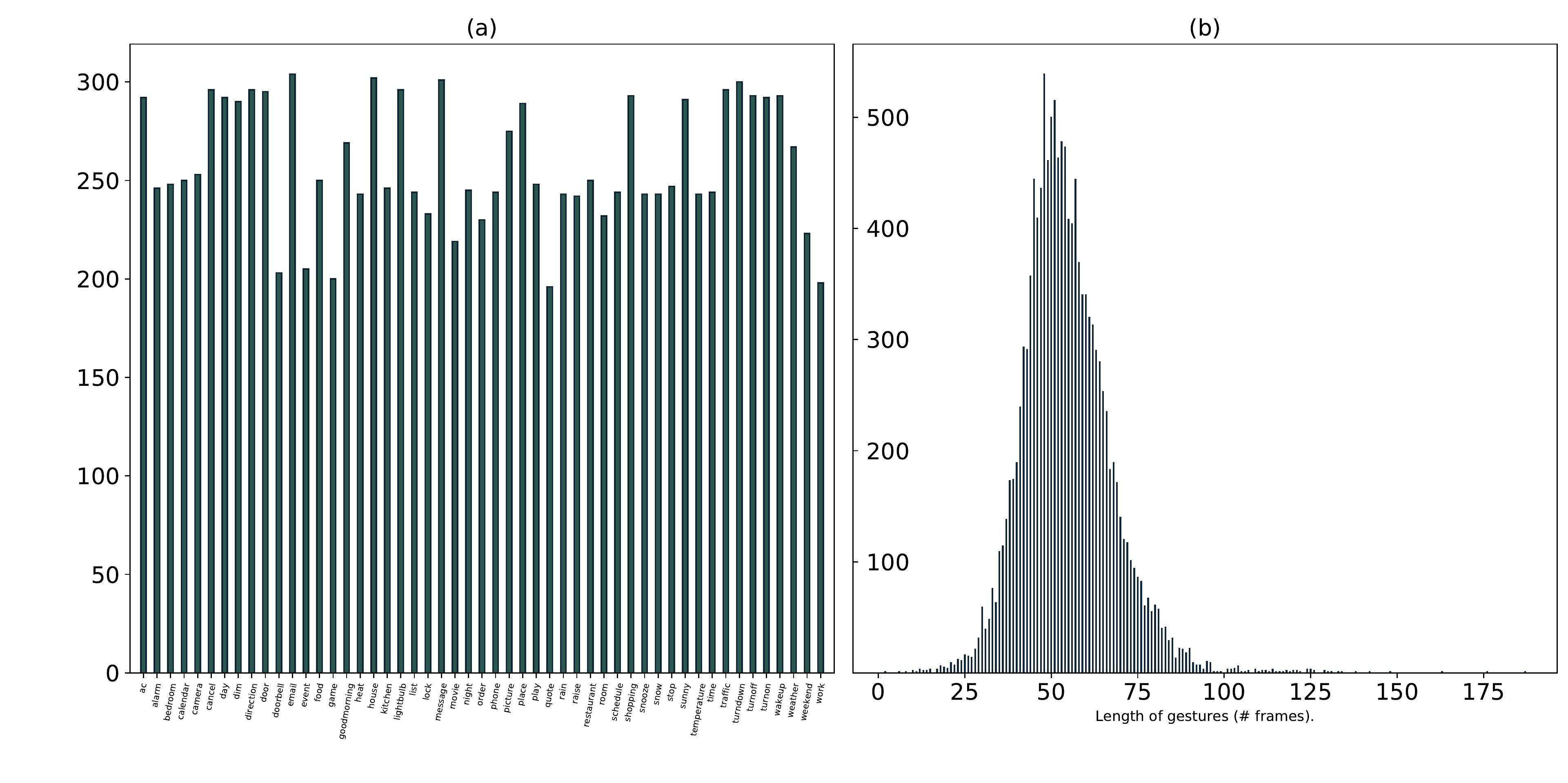}
	\end{center}
	\caption{a) Distribution of number of samples per gesture class. b) Distribution of gesture sample duration; X-axis represents gesture length in terms of number of frames in a video.}
	\label{fig:class_hist}
    \end{figure*}
	\subsection{Collection Protocol}
	The data was collected with a Microsoft Kinect version 2.0 depth camera positioned in front of the signer. For each sign (a single class like \texttt{Air Condition} or \texttt{AC}) we collected 24 samples continuously; and the process was repeated for every sign (51 classes in total). Due to time and availability constraints, for some subjects we could not collect the samples for all the classes resulting in a total of 13107 samples. The distance between the subject and the sensor was varied in the range from 10 to 15 feet to simulate practical scenarios. No constraints were imposed on performers' posture and lighting condition of the room.

To gather individual samples from the continuous data, segmentation marks were interleaved through a user interface. This was later used to segment individual samples. These samples were further segmented using motion calculation of the wrist joint co-ordinates from skeletal data. Figure \ref{fig:class_hist} (a) illustrates the distribution of number of samples per gesture class in GMU-ASL51 dataset. Figure \ref{fig:class_hist} (b) shows the distribution of duration of videos in our dataset.
	
	\subsection{Data Modality}
	
\begin{figure}[h]
		\begin{center}
			\includegraphics[width=.9\linewidth]{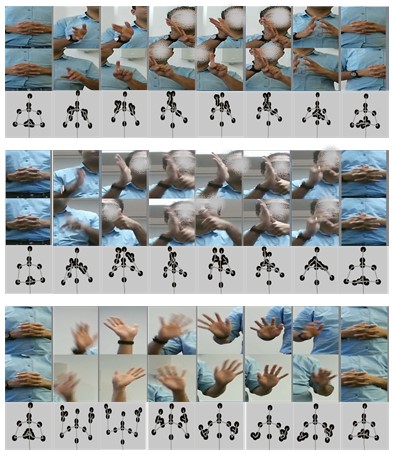}
		\end{center}
		\caption{Visualization of hand shapes and skeletal joints of two sign classes. Top panel shows the sign \texttt{Alarm} and middle panel shows the sign \texttt{Doorbell}. For each sign, first two rows are the left and right hand image patches and third row is the skeletal configuration. We can see for \texttt{Alarm} and \texttt{Doorbell}, the skeletal motion is almost similar but has different hand shapes. Bottom panel shows another sign \texttt{Weather} which has quite distinguishable skeletal motion from top two.}
		\label{fig:alarm_doorbell_conf}
	\end{figure}

	All of our experiments on ASL recognition were done with RGB video data and/or skeletal data. Skeletal data is a multivariate, multidimensional time series input where each body part acts as a variable and each of them have 3D coordinate data at each time step. The skeletal data provides motion trajectory of different body parts such as wrist, elbow and shoulder (total 25 such body parts) over whole video frames. This process is called skeletal tracking.
	Skeletal data provides high level motion of different body parts. These are useful for capturing discriminant features associated with different types of gestures. 
	However, for better modeling of sign language, hand shape is crucial, as different signs may have similar motion but different hand shapes and orientation.
	Figure \ref{fig:alarm_doorbell_conf} presents one such example where the sign pair \texttt{Alarm} and \texttt{Doorbell} have exact same motion pattern according to skeletal data but have different hand shapes. We observe similar situation for sign pairs such as \texttt{Kitchen}/\texttt{Room}, \texttt{Time}/\texttt{Movie}, \texttt{Quote}/\texttt{Camera}, \texttt{Lock}/\texttt{Stop} and many more. 
	
	We hypothesize  that hand shape is  useful in situations where skeletal data has similar dynamic motion pattern for different sign classes. Due to this fact, we extract and use hand shape patterns from RGB video data. 
	
\section{Our Approach}
	Inspired by the success of deep learning approaches in computer vision \cite{NIPS2012_4824}, we applied different deep learning architectures to model sign languages from both input modes (RGB and skeletal). Unlike traditional image classification or object detection models where neural networks learn hierarchical spatial features from data, sign recognition requires capture of temporal body motion.

	\subsection{Recurrent Neural Networks (RNN)}
\begin{figure}[h]
		\begin{center}
			\includegraphics[clip, trim=6cm 25cm 11cm 4cm,width=\linewidth]{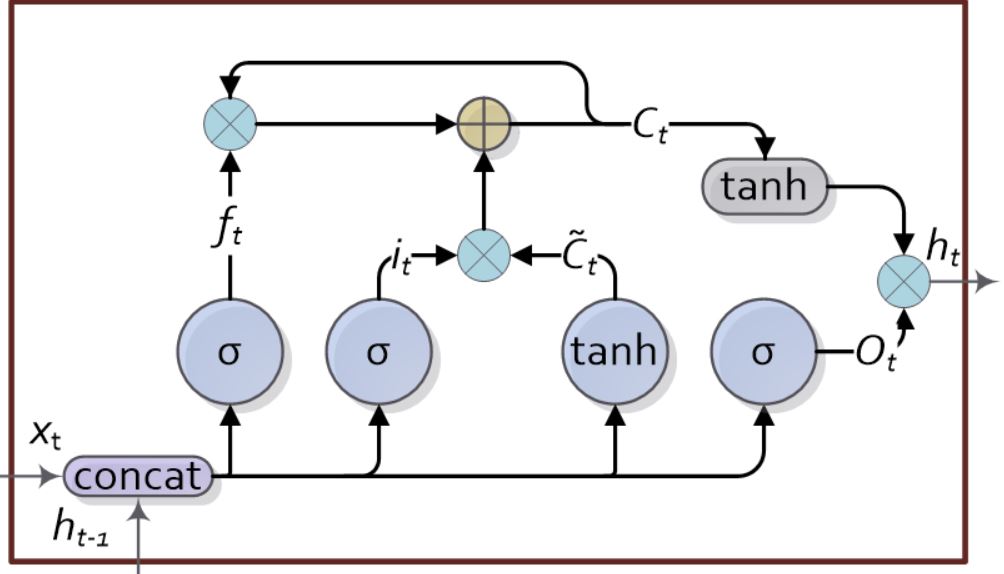}
		\end{center}
		\caption{An LSTM cell. Four circles represent four different neural network which act as gates.}
		\label{fig:lstm_cell}
	\end{figure}

	RNN has shown success in modeling sequential pattern in data\cite{DBLP:journals/corr/Lipton15}. It can capture temporal dynamics in data by maintaining an internal state. 
	However, the basic RNN has problems dealing with long term dependencies in data due to the vanishing gradient problem \cite{298725}. 
	Some solutions to the vanishing gradient problem involve 
	careful initialization of network parameters or  early stopping  \cite{Bengio:1994:LLD:2325857.2328340}. But the most effective solution is 
	to modify the RNN architecture in such a way that there exists a memory state (cell state) at every time step that can identify what to remember and what to forget. This architecture is referred to as
	long short term memory (LSTM) network \cite{Hochreiter:1997:LSM:1246443.1246450}. 
	While the basic RNN is a 
	direct transformation of the previous state and the current input, the 
	LSTM maintains an internal memory and  has
	mechanisms to update and use that memory. This is achieved by deploying four separate neural networks also 
	called gates. Figure~\ref{fig:lstm_cell} depicts a cell of an LSTM network which shows input at the current time step ${x_t}$ and the 
	previous state ${h_{t-1}}$ enter into the cell; and get concatenated. 
	The forget gate processes it to remove
	unnecessary information, and outputs ${f_t}$ which gets multiplied with the previously 
	stored memory ${C_{t-1}}$ and produces a refined memory for the current time. 
	
	\begin{equation}
	\label{eq:lstm_eq}
	\begin{aligned}
	{f_t} &= \sigma(W_f \times concat({h_{t-1}}, {x_t})) \\
	{i_t} &= \sigma(W_i \times concat({h_{t-1}}, {x_t})) \\
	{\tilde{C}_t} &= tanh(W_{\tilde{C}} \times concat({h_{t-1}}, {x_t})) \\
	{C_t} &= ({f_t} \otimes {C_{t-1}}) \oplus ({i_t} \otimes {\tilde{C}_t)}  \\
	{o_t} &= \sigma(W_o \times concat({h_{t-1}}, {x_t})) \\
	{h_t} &= {o_t} \otimes tanh({C_t})
	\end{aligned}
	\end{equation}
    \begin{figure}
		\begin{center}
			\includegraphics[clip, trim= 10.0cm 9.0cm 17.0cm 4.0cm, width=\linewidth]{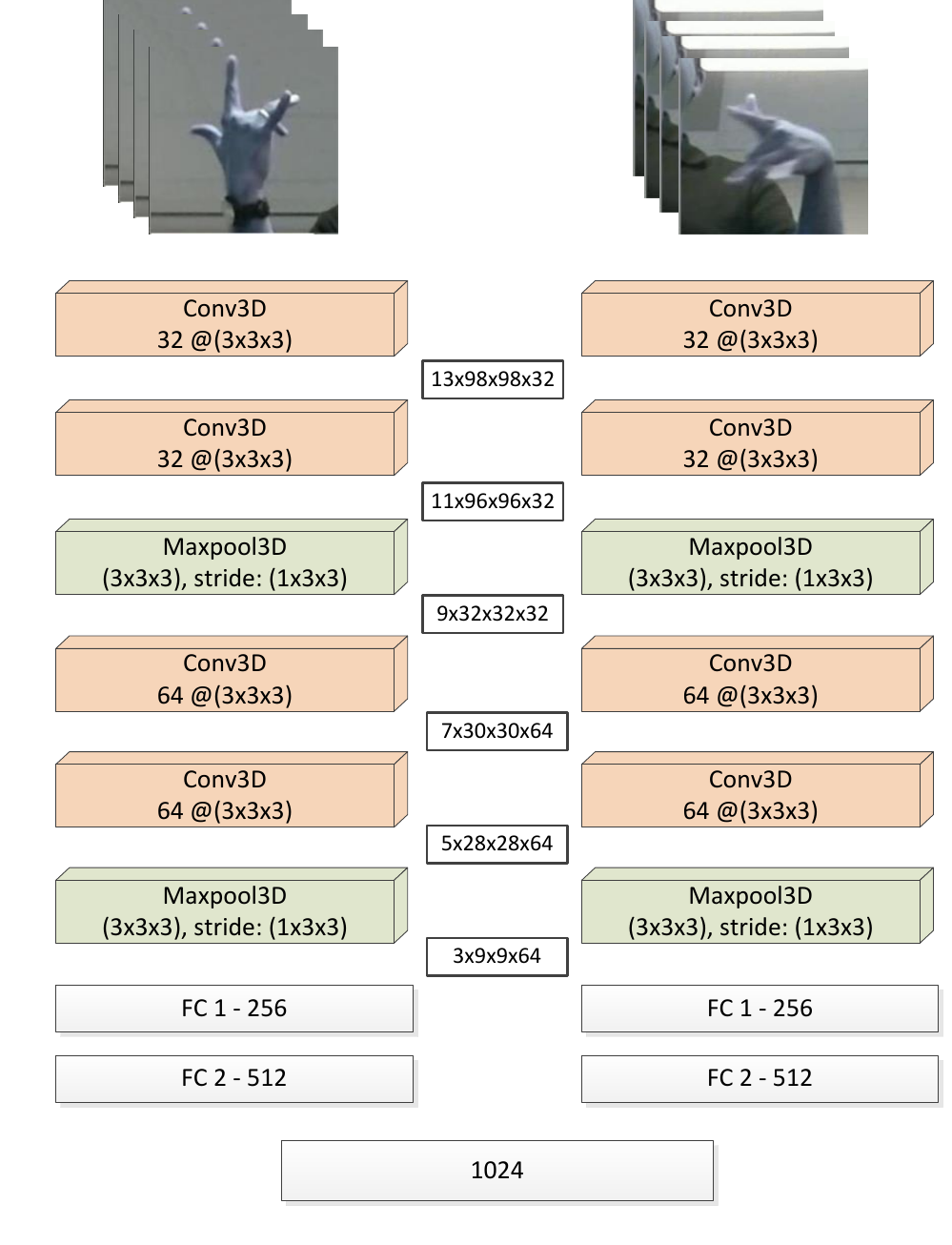}
		\end{center}
		\caption{Used 3D CNN architecture for this work. It consists of four 3D convolutional layers and two fully connected layers at the end. There are two separate networks for left and right hands. Final embedding of these two networks are concatenated before producing softmax score. Feature map dimensions after each layer are shown in the middle.}
		\label{fig:cnn_arch}
	\end{figure}
	Meanwhile, the input 
	and  update gate process the concatenated input and convert it into a candidate memory for the current time step by element--wise multiplication. The refined memory and proposed candidate memory of the current step are added to produce the final memory for the current step. This addition could render the output to be out of scale. To avoid that, a squashing function (hyperbolic tan) is used, which scales the elements of the output vector into a fixed range. Finally ${o_t}$, the output 
	from output gate gets multiplied with the squashing function and produces the current time step output. Figure \ref{fig:lstm_cell} shows an LSTM cell.
	The forget, input, update and output gates are represented by four circles and symbolized as $f_t$, $i_t$, $\tilde{C_t}$ and $o_t$, respectively . Equation \ref{eq:lstm_eq} shows LSTM functions; where $\oplus$ and $\otimes$ represent element wise addition and multiplication respectively; $\times$ represents matrix multiplication, $concat$ process means a concatenation of its input. 
	
	\begin{figure*}[h]
		\begin{center}
			\includegraphics[clip, trim = 16.0cm 19.0cm 34.0cm 6.0cm, width=\linewidth]{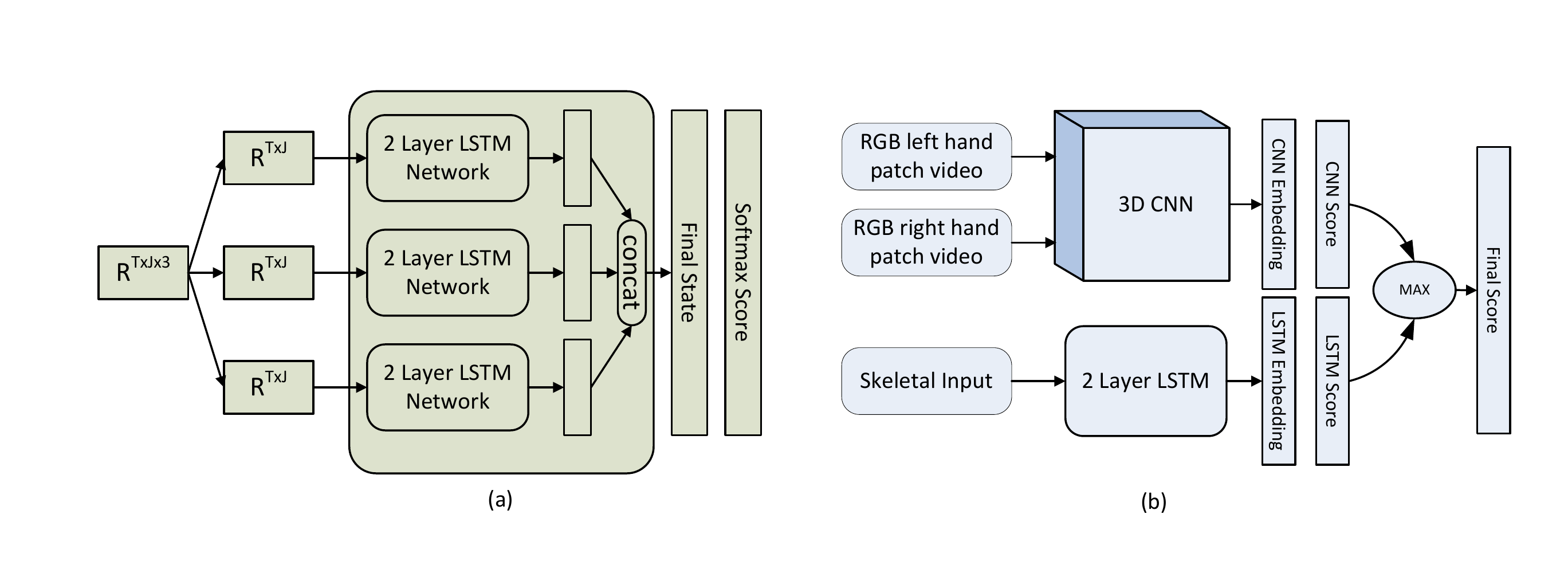}
		\end{center}
		\caption{Proposed architectures. Fig (a): Axis independent LSTM network where data from each axis enters into different LSTM networks and at the end we take the  concatenation of individual states. Fig (b): Combined architecture. Here 3D CNN symbolizes the architecture we presented in Figure \ref{fig:cnn_arch}. Here both CNN and LSTM network model data separately. At the end we take the maximum of probability scores produced by both network.}
		\label{fig:architectures}
	\end{figure*}
	\subsection{3D Convolutional Neural Network}
	Traditional convolutional neural network (CNN) is two dimensional in which each layer has a stack of 2D feature maps generated from previous layer or from inputs in case of first layer. A layer also has a certain numbers of filters which are rectangular patches of parameters. Each filter convolves over the stack of 2D feature maps at previous layer and produces feature maps (equal to the number of filters in the current layer) at current layer. The operation is given by Equation \ref{2d_conv} where $F_{i,j}^{l}$ denotes the value of feature map at $l^{th}$ layer at location $(i,j)$. $\odot$ represents dot product of filter $W$ and associated feature map patch in previous layer.
	\begin{align}
	\label{2d_conv}
	F_{i,j}^{l} =& \sum_{m} W_{i,j} \odot F_{i,j}^{l-1} 
	\end{align}
	Standard CNN fails to capture the temporal information associated with data, which is important in video or any type of sequential data representation. To solve this problem, 3D convolution was introduced in \cite{Ji:2013:CNN:2412386.2412939}. The key difference is that kernels are 3D and sub sampling (pooling) layers work across three dimensions. 
	\begin{align}
	\label{3d_conv}
	F_{i,j,k}^{l} =& \sum_{m} W_{i,j,k} \odot F_{i,j,k}^{l-1}
	\end{align}%
	Equation \ref{3d_conv} shows 3D convolution function. In this case from each filter we get a 3D feature map and $F_{i,j,k}$ denotes value at  $(i, j,k)$ location after convolution operation. The dot product is between two  three-dimensional matrices (also called tensors). 
	
	

 	\subsection{Axis Independent LSTM}
 	\label{axis_ind_rnn}
 	Given a sample skeletal data of $R^{T \times J \times 3}$, where $T$ denotes time axis, $J$ is the number of body joints and the last dimension is the 3D coordinates of each joint. We flatten every dimension except time and at each time step we can feed a vector of size $R^{3 \times J}$ as input. However, we have empirically verified that learning a sequential pattern for each coordinate axis independently and combining them later shows stronger classification performance. Based on this, we trained three different 2 layer LSTMs for data from x, y, and z coordinates separately; and concatenate their final embedding to produce softmax output. In this setting, each separate LSTM receives data as $R^{T \times J}$ and final embedding size is $R^{3\times S}$ where $S$ is the state size of LSTM cell. Figure \ref{fig:architectures} (a) shows the architecture where as a sample arrives, just before entering into main network, data along separate axis is split and entered into three different LSTM networks. The model concatenates the final state from each of the separate LSTM networks; followed by feeding this into the softmax layer for classification. This approach is referred by Axis Independent Architecture (AI-LSTM). Implementation details such as values of T and J are provided in the `Experiments' section.
 	
 	\subsection{Spatial AI-LSTM}
\begin{figure}[h]
 		\begin{center}
 			\includegraphics[width=\linewidth]{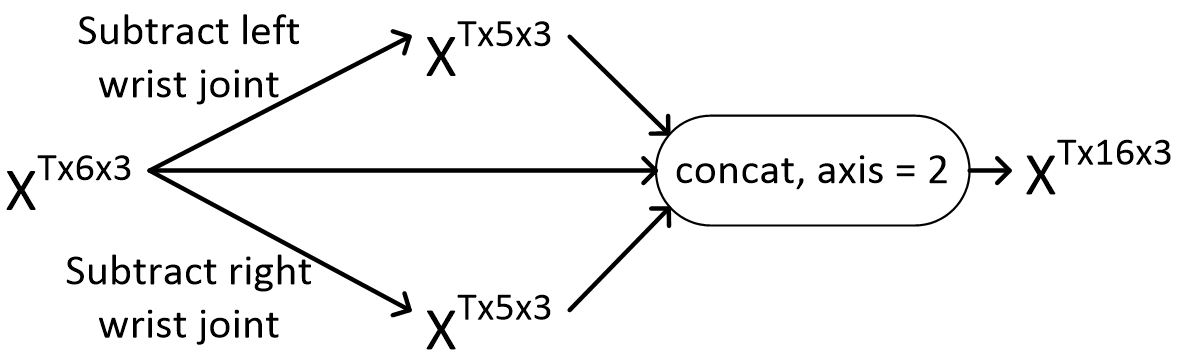}
 		\end{center}
 		\caption{Spatial data augmentation.}
 		\label{fig:data_aug}
 	\end{figure}

 	AI-LSTM, described in last section, works by modeling temporal dynamics of body joints' data over time. However, there can be spatial interactions with joints at a specific time step. 
 	It fails to capture any such interaction among joints in a given time.
 	To incorporate spatial relationship among joints, we propose a simple novel data augmentation technique for skeletal data. We do this by origin transfer. For each frame in a gesture sample, we use each wrist joints as origin and transform all other joints' data by subtracting that origin from them. In this way spatial information is added to the input. We refer this model with spatial data augmentation as Spatial AI-LSTM. This augmentation technique is depicted in Figure \ref{fig:data_aug}.
 	A sample data of form $R^{T \times 6 \times 3}$ results in a representation of $R^{T \times 5 \times 3}$ after subtracting left wrist joint (origin transfer).
 	After this augmentation process, each sample is a  $R^{20 \times 16 \times 3}$ matrix.  Hence, each separate LSTM networks in our Spatial AI-LSTM network receives an input of $R^{20 \times 16}$.
 		
 \begin{figure*}[h]
	\begin{center}
		\includegraphics[width=0.8\linewidth]{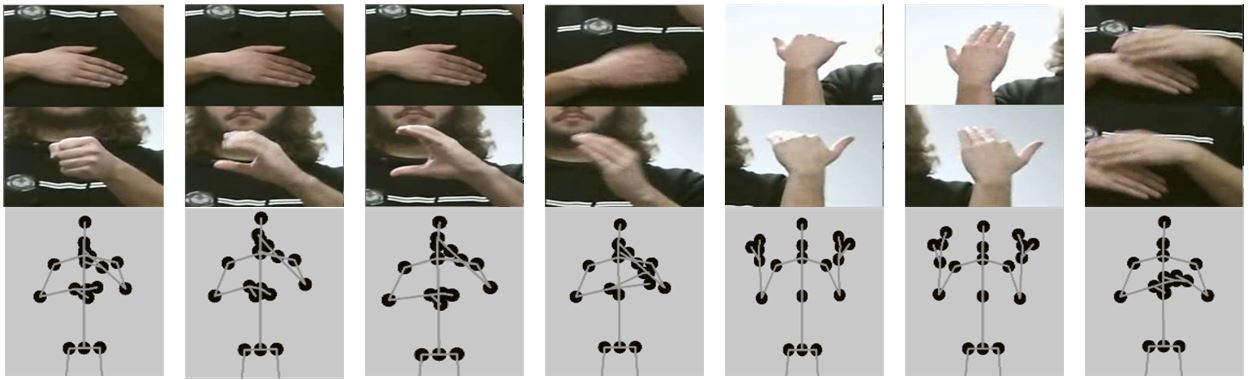}
	\end{center}
	\caption{Seven sampled frames from a sign of class Air Condition. Top two panels show cropped hand patches while bottom panel shows skeletal configuration of corresponding frames.}
	\label{fig:ac_sign_sub1}
\end{figure*}
 	\subsection{Combined Network}
 	We hypothesize that, some signs that have mostly similar skeletal motion pattern could be distinguishable using hand shape information.  We propose a combination of LSTM and 3D CNN networks. We call this Max CNN-LSTM network. Figure \ref{fig:architectures} (b) represents the the Max CNN-LSTM. The details of 3D CNN module is shown in Figure \ref{fig:cnn_arch}.
 	This architecture has two parts: one for left hand patches and other for right hand patches. Each part has four 3D convolutional layers (second and fourth layers have following maximum pooling layers) followed by 2 fully connected layers.  Final embeddings from these two parts are concatenated and by using a softmax layer, from which a classification score is produced. The other AI-LSTM network is fed with skeletal time series data. At the final time step, the LSTM state vector is taken and using a softmax layer another probability score is produced. The final classification score is created by taking element wise maximum of the output scores from the two networks. During back--propagation, both networks are trained on their own score. The combined network acts like a model ensemble and some sign classes which are confused by RNN network alone might have an improved recognition accuracy with this approach.

	
	\section{Experiments}
	
		 Naturally each sign has different frame length after segmentation because each subject does a sign at different speed. It is possible that the same subject may do the same sign at different speeds at different times which makes the recognition challenging. Further, neighboring frames contain redundant information; and all joints will not have equal amount of motion or pattern in case of skeletal data. 
	 
	 
	 
	 \subsection{Skeletal Data}
	 Most of the signs do not involve all the 25 joints' information provided by Kinect sensor; specifically, joints involved with the two hands convey most information.  Based on this, we consider only 6 joints (wrist, elbow, shoulder) from both as input to the LSTM network.  Figure \ref{fig:ac_sign_sub1} shows an example  where 7 frames were sampled from a sign video of class \texttt{Air Condition} and the   bottom panel shows the skeletal configuration across those 7 frames. From each sign video we sampled some number of frames uniformly and took joints' data  associated with those frames. We verified empirically that picking 20 frames uniformly works best for skeletal data. For samples with less than 20 samples we convert them to 20 frame signs by interleaving existing frames uniformly. Thus skeletal data for each sample is a vector in $R^{20 \times 6 \times 3}$.
	 

 	\subsection{Video Data}
 	Since ASL involves specific hand shape patterns, we crop both hand regions at each frame. 
 	Using 2D coordinates of hand joints on a video frame as center, we do a $100 \times100$ crop to generate hand patches.
 	To reduce motion blur, we calculate velocity of joints at each video frame using skeletal coordinates and then sample from frames which have less motion. 
	We sampled 15 frames from each sign video resulting in a vector of $R^{15 \times 100 \times 100 \times 3}$ for each hand patch.

	\subsection{Training Details}
	To deal with over-fitting, dropout was used for all networks except convolutional layers with probability of 0.5. In addition to dropout, L2 regularization was used for LSTM networks and for dense layers; $\beta$ was set to 0.008 which controls the impact of regularization on the network. State size and number of layers of LSTM networks were 50 and 2, respectively. Learning rate for Max CNN-LSTM and LSTM networks were set to  $0.00001$ and $0.00005$, respectively. We used Adam Optimizer for training our networks~\cite{DBLP:journals/corr/KingmaB14}. All networks were run for a certain number of epochs (200-300) with a batch size of 64. We developed all of our models with Tensorflow 1.10 (python). 
	Average time taken to train an AI-LSTM and an Spatial AI-LSTM are 25 and 30 minutes on an Intel(R) Core(TM) i5-7600 (3.50GHz) processor respectively.
    We trained 3D CNN and Max 3D CNN models on GPU (Tesla K80) and each model took around 20 hours to train.

	\subsection{Baseline Methods}
	We use support vector machines and random forest for baseline comparison. The baseline models utilize skeletal data in each axis for every joint in building the following features per sample: Mean, Area, Skew, Kurtosis, Motion Energy, Range and Variance over the frames \cite{baseline_feature_based}. We have 6 upper body joints and 3 axes per joint and 7 features for each giving a total of 126 $(7 \times 6 \times 3)$ features per sample.
	
	\begin{table}[h]
	\caption{Average cross subject (CS) accuracy across all test subjects for different proposed architectures and baselines. Standard deviation across test subjects' accuracy is also shown.}
		\centering
		\begin{tabular}{lcc}
			\hline
			Methods  &  Accuracy (CS)  & Std. Deviation\\
			\hline
			SVM    & 62\%  &  10\% \\
			Random Forest   & 66\% & 8\% \\
			3D CNN       & 52\% & 12\%\\
			AI-LSTM     & 73\% & 6\% \\
			Max CNN-LSTM & \textbf{75\%} & 7\% \\
			Spatial AI-LSTM & \textbf{81\%} & 6\% \\
			\hline
		\end{tabular}
		\label{tab:result_table}
	\end{table}
	\subsection{Experimental Results}
	\label{sec_results}
	Table \ref{tab:result_table} shows the comparative results among our proposed architectures and baselines. 
	Overall, we use data from 12 subjects for our experiments which sum up to 13107 sign gesture samples in total. To evaluate model performance on a specific subject (test subject), we adopt cross subject evaluation criteria. Suppose, X is the test subject. We train our networks with all sign samples except those are from subject X. We use subject X's data as test split to evaluate the performance of the networks. Table \ref{tab:result_table} shows the average test accuracy for all 12 subjects. We can see that 3D CNN network alone performs  worse than simpler baselines. But when coupled with AI-LSTM as Max CNN-LSTM, it shows an  increase in recognition accuracy by 2\% from AI-LSTM alone. This is because some of the signs  are confused by the AI-LSTM network because of similar skeletal motion pattern. 
	Incorporating spatial relationship among joints leads to a 
	significant accuracy gain. The Spatial AI-LSTM is trained only on skeletal data but outperforms the combined network by 6\%. 
	\begin{figure}[h]
		\begin{center}
			\includegraphics[clip, trim= .5cm 0.0cm 0.5cm 0.0cm, width=\linewidth, height=.6\textheight]{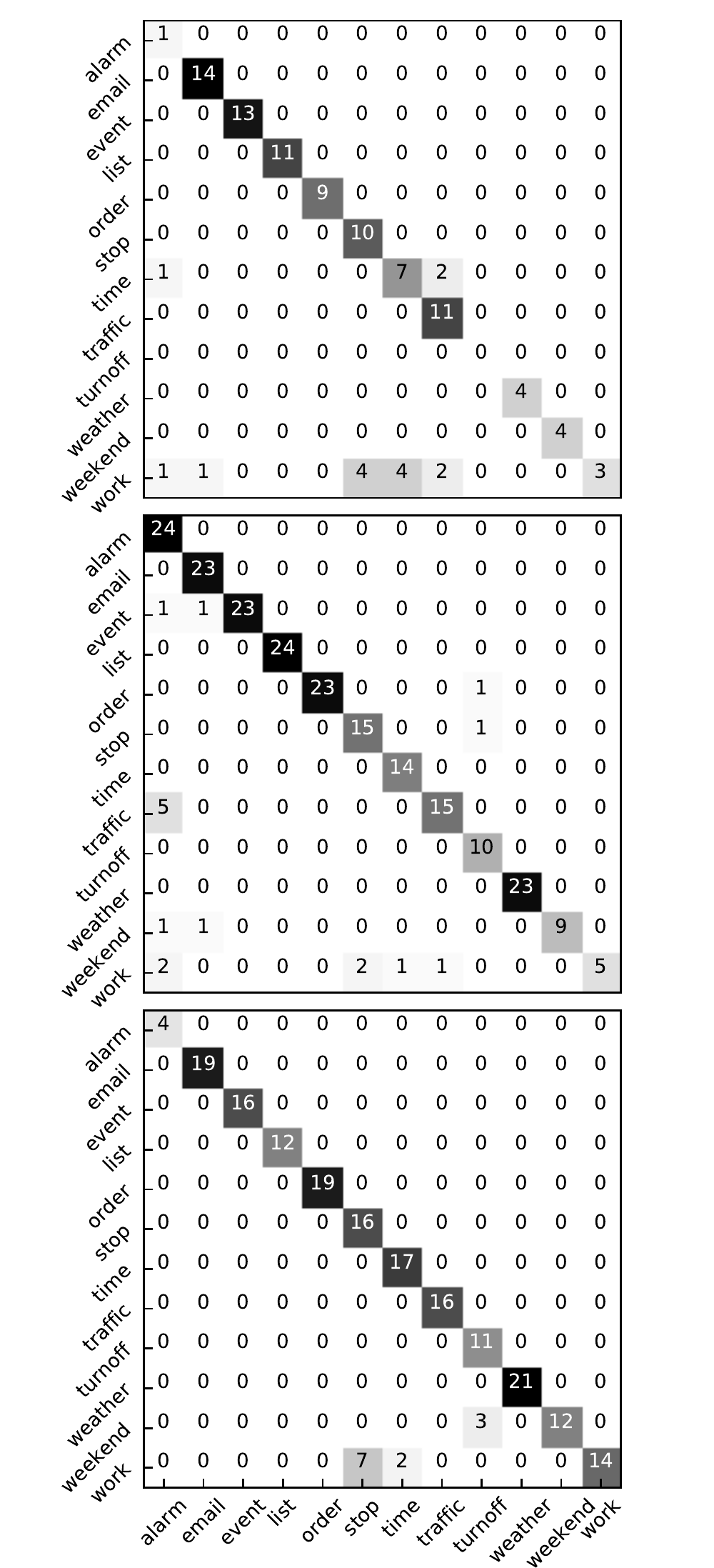}
		\end{center}
		\caption{Confusion matrix for a subset of sign classes from a subject for AI-LSTM, Max CNN-LSTM and Spatial AI-LSTM from top to bottom respectively. Mentioned signs are a subset of 51 sign classes.}
		\label{fig:cm_subset}
	\end{figure} 
	
	Figure \ref{fig:cm_subset} shows three confusion matrices for a subset of twelve sign classes for a subject. The top matrix is for AI-LSTM network, middle one is for  Max CNN-LSTM and bottom one is for Spatial AI-LSTM. As seen  in Figure \ref{fig:alarm_doorbell_conf} the sign pairs \texttt{Alarm}/\texttt{Doorbell} are similar in skeletal motion but have different hand shapes. Since Max CNN-LSTM includes hand shapes, it can successfully recognize it while other two models struggles. Same is true for some other signs like \texttt{Email}, \texttt{Event}, \texttt{List}, \texttt{Order} and \texttt{Weather} . Some other signs are better recognized by Spatial AI-LSTM network. It should be mentioned here that accuracy listed in Table \ref{tab:result_table} shows average accuracy across all test subjects, while Figure \ref{fig:cm_subset} presents confusion matrix for a single test subject. For this particular subject overall test accuracy is 58\%, 70\% and 69\% for AI-LSTM, Max CNN-LSTM and Spatial AI-LSTM network respectively. 

%
    \subsection{Effect of Same Subject Data in Training}	
	In addition to having the cross subject accuracy described in section \ref{sec_results}, we also want to know the impact of adding a test subject's data to the training process. It is obvious that adding test subject's data to the training must increase the accuracy of the network for the subject. However, we want to know how much or what fraction of data is necessary for significant improvement in performance.
	This is important for assessing the practial usability of a recognition system. In other words, we want to know how quickly or with what amount of data, the current system can be adapted for a subject completely unknown to the system. To do that, we first pick a test subject and train a model for the test subject with data from all other subjects in our dataset. Then we retrain the model with some fraction of data from the test subject. We keep increasing the fraction of data being used from the test subject in the retraining process up to $50\%$. The other half of the test subject's data is used for testing the model.
	
	\begin{figure}[h]
		\begin{center}
			\includegraphics[clip, trim= 0.0cm 0.0cm 0.2cm 0.2cm, width=\linewidth]{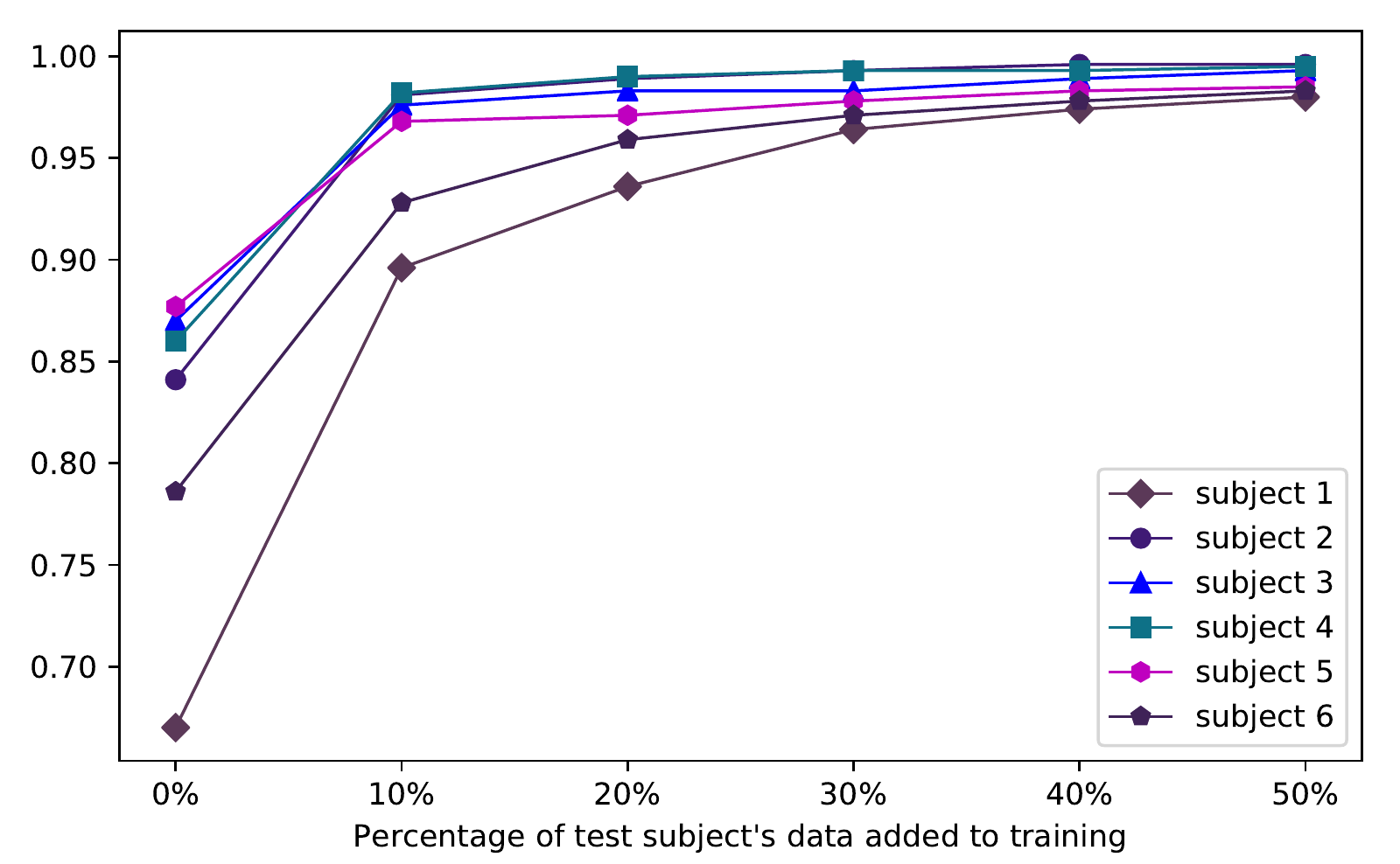}
		\end{center}
		\caption{Effect of adding data to the training from test subject in Spatial AI-LSTM model. X axis is the fraction of test subject's data used in training. Y axis is the test accuracy.}
		\label{fig:train_p}
	\end{figure}
	
	Figure \ref{fig:train_p} shows the effect of added training data from test subjects in the retraining on six subjects from our dataset in case of Spatial AI-LSTM model. We see that, adding data from a test subject increase recognition accuracy for all of the subjects shown. It is interesting to observe that adding even $10\%$ of data from a test subject gives significant improvement in recognition accuracy (close to $95\%$) for almost all of the subjects shown. 
	
	\section{Conclusion}
	
	We present a deep learning based approach for ASL recognition that leverages skeletal and video data. The proposed model captures the underlying temporal dynamics associated with sign language and also identifies specific hand shape patterns from video data to improve recognition performance. 
	A new data augmentation technique was introduced that allowed the LSTM networks to capture spatial dynamics among joints. %
	Finally, a large public dataset for ASL recognition will be released to the community to spur research in this direction; and bring benefits of digital assistants to the deaf and hard of hearing community. For future research direction, we are looking into the problem of sentence level ASL recognition. We also plan to use other data modality such as wifi signals which can be complimentary to video in sign language recognition.

	\section*{Acknowledgments}
	This work was supported by Google Research Award.
    Some of the experiments were run on ARGO, a research computing cluster provided by the Office of Research Computing at George Mason University, VA. (URL:http://orc.gmu.edu)	
	\bibliographystyle{IEEEtran}
	\bibliography{IEEEabrv,dsaa_submission}

\begin{thebibliography}{10}
\providecommand{\url}[1]{#1}
\csname url@samestyle\endcsname
\providecommand{\newblock}{\relax}
\providecommand{\bibinfo}[2]{#2}
\providecommand{\BIBentrySTDinterwordspacing}{\spaceskip=0pt\relax}
\providecommand{\BIBentryALTinterwordstretchfactor}{4}
\providecommand{\BIBentryALTinterwordspacing}{\spaceskip=\fontdimen2\font plus
\BIBentryALTinterwordstretchfactor\fontdimen3\font minus
  \fontdimen4\font\relax}
\providecommand{\BIBforeignlanguage}[2]{{%
\expandafter\ifx\csname l@#1\endcsname\relax
\typeout{** WARNING: IEEEtran.bst: No hyphenation pattern has been}%
\typeout{** loaded for the language `#1'. Using the pattern for}%
\typeout{** the default language instead.}%
\else
\language=\csname l@#1\endcsname
\fi
#2}}
\providecommand{\BIBdecl}{\relax}
\BIBdecl

\bibitem{3072291}
P.~Kushalnagar, G.~Mathur, C.~J. Moreland \emph{et~al.}, ``Infants and children
  with hearing loss need early language access,'' \emph{The Journal of clinical
  ethics}, pp. 1--1, 2010.

\bibitem{7177428}
J.~Huang, W.~Zhou, H.~Li, and W.~Li, ``Sign language recognition using 3d
  convolutional neural networks,'' in \emph{2015 IEEE International Conference
  on Multimedia and Expo (ICME)}, June 2015, pp. 1--6.

\bibitem{Ji:2013:CNN:2412386.2412939}
\BIBentryALTinterwordspacing
S.~Ji, W.~Xu, M.~Yang, and K.~Yu, ``3d convolutional neural networks for human
  action recognition,'' \emph{IEEE Trans. Pattern Anal. Mach. Intell.},
  vol.~35, no.~1, pp. 221--231, Jan. 2013. [Online]. Available:
  \url{http://dx.doi.org/10.1109/TPAMI.2012.59}
\BIBentrySTDinterwordspacing

\bibitem{Sun:2015:LSV:2753829.2629481}
\BIBentryALTinterwordspacing
C.~Sun, T.~Zhang, and C.~Xu, ``Latent support vector machine modeling for sign
  language recognition with kinect,'' \emph{ACM Trans. Intell. Syst. Technol.},
  vol.~6, no.~2, pp. 20:1--20:20, Mar. 2015. [Online]. Available:
  \url{http://doi.acm.org/10.1145/2629481}
\BIBentrySTDinterwordspacing

\bibitem{Zhang:2012:MKS:2225053.2225203}
\BIBentryALTinterwordspacing
Z.~Zhang, ``Microsoft kinect sensor and its effect,'' \emph{IEEE MultiMedia},
  vol.~19, no.~2, pp. 4--10, Apr. 2012. [Online]. Available:
  \url{http://dx.doi.org/10.1109/MMUL.2012.24}
\BIBentrySTDinterwordspacing

\bibitem{7298714}
Y.~Du, W.~Wang, and L.~Wang, ``Hierarchical recurrent neural network for
  skeleton based action recognition,'' in \emph{2015 IEEE Conference on
  Computer Vision and Pattern Recognition (CVPR)}, June 2015, pp. 1110--1118.

\bibitem{DBLP:journals/corr/Lipton15}
\BIBentryALTinterwordspacing
Z.~C. Lipton, ``A critical review of recurrent neural networks for sequence
  learning,'' \emph{CoRR}, vol. abs/1506.00019, 2015. [Online]. Available:
  \url{http://arxiv.org/abs/1506.00019}
\BIBentrySTDinterwordspacing

\bibitem{copycat_zafrulla}
Z.~{Zafrulla}, H.~{Brashear}, P.~{Yin}, P.~{Presti}, T.~{Starner}, and
  H.~{Hamilton}, ``American sign language phrase verification in an educational
  game for deaf children,'' in \emph{2010 20th International Conference on
  Pattern Recognition}, Aug 2010, pp. 3846--3849.

\bibitem{Zafrulla:2011:ASL:2070481.2070532}
\BIBentryALTinterwordspacing
Z.~Zafrulla, H.~Brashear, T.~Starner, H.~Hamilton, and P.~Presti, ``American
  sign language recognition with the kinect,'' in \emph{Proceedings of the 13th
  International Conference on Multimodal Interfaces}, ser. ICMI '11.\hskip 1em
  plus 0.5em minus 0.4em\relax New York, NY, USA: ACM, 2011, pp. 279--286.
  [Online]. Available: \url{http://doi.acm.org/10.1145/2070481.2070532}
\BIBentrySTDinterwordspacing

\bibitem{978-3-319-16178-5_40}
L.~Pigou, S.~Dieleman, P.-J. Kindermans, and B.~Schrauwen, ``Sign language
  recognition using convolutional neural networks,'' in \emph{Computer Vision -
  ECCV 2014 Workshops}, L.~Agapito, M.~M. Bronstein, and C.~Rother, Eds.\hskip
  1em plus 0.5em minus 0.4em\relax Cham: Springer International Publishing,
  2015, pp. 572--578.

\bibitem{ZAKI2011572}
\BIBentryALTinterwordspacing
M.~M. Zaki and S.~I. Shaheen, ``Sign language recognition using a combination
  of new vision based features,'' \emph{Pattern Recognition Letters}, vol.~32,
  no.~4, pp. 572 -- 577, 2011. [Online]. Available:
  \url{http://www.sciencedirect.com/science/article/pii/S016786551000379X}
\BIBentrySTDinterwordspacing

\bibitem{Cooper2017}
H.~Cooper, E.-J. Ong, N.~Pugeault, and R.~Bowden, ``Sign language recognition
  using sub-units", booktitle="gesture recognition,'' S.~Escalera, I.~Guyon,
  and V.~Athitsos, Eds.\hskip 1em plus 0.5em minus 0.4em\relax Cham: Springer
  International Publishing, 2017, pp. 89--118.

\bibitem{7780484}
A.~Shahroudy, J.~Liu, T.~T. Ng, and G.~Wang, ``Ntu rgb+d: A large scale dataset
  for 3d human activity analysis,'' in \emph{2016 IEEE Conference on Computer
  Vision and Pattern Recognition (CVPR)}, June 2016, pp. 1010--1019.

\bibitem{8101019}
J.~Liu, A.~Shahroudy, D.~Xu, A.~K. Chichung, and G.~Wang, ``Skeleton-based
  action recognition using spatio-temporal lstm network with trust gates,''
  \emph{IEEE Transactions on Pattern Analysis and Machine Intelligence}, pp.
  1--1, 2017.

\bibitem{7410817}
V.~Veeriah, N.~Zhuang, and G.~J. Qi, ``Differential recurrent neural networks
  for action recognition,'' in \emph{2015 IEEE International Conference on
  Computer Vision (ICCV)}, Dec 2015, pp. 4041--4049.

\bibitem{song2016end}
S.~Song, C.~Lan, J.~Xing, W.~Zeng, and J.~Liu, ``An end-to-end spatio-temporal
  attention model for human action recognition from skeleton data,'' in
  \emph{AAAI Conference on Artificial Intelligence}, 2017, pp. 4263--4270.

\bibitem{8226767}
J.~Liu, G.~Wang, L.~Y. Duan, K.~Abdiyeva, and A.~C. Kot, ``Skeleton-based human
  action recognition with global context-aware attention lstm networks,''
  \emph{IEEE Transactions on Image Processing}, vol.~27, no.~4, pp. 1586--1599,
  April 2018.

\bibitem{DBLP:journals/corr/abs-1711-05941}
\BIBentryALTinterwordspacing
J.~Liu, N.~Akhtar, and A.~Mian, ``Skepxels: Spatio-temporal image
  representation of human skeleton joints for action recognition,''
  \emph{CoRR}, vol. abs/1711.05941, 2017. [Online]. Available:
  \url{http://arxiv.org/abs/1711.05941}
\BIBentrySTDinterwordspacing

\bibitem{DBLP:conf/cvpr/KeBASB17}
\BIBentryALTinterwordspacing
Q.~Ke, M.~Bennamoun, S.~An, F.~A. Sohel, and F.~Boussa{\"{\i}}d, ``A new
  representation of skeleton sequences for 3d action recognition,'' in
  \emph{2017 {IEEE} Conference on Computer Vision and Pattern Recognition,
  {CVPR} 2017, Honolulu, HI, USA, July 21-26, 2017}, 2017, pp. 4570--4579.
  [Online]. Available: \url{https://doi.org/10.1109/CVPR.2017.486}
\BIBentrySTDinterwordspacing

\bibitem{4563181}
V.~{Athitsos}, C.~{Neidle}, S.~{Sclaroff}, J.~{Nash}, A.~{Stefan}, , and
  A.~{Thangali}, ``The american sign language lexicon video dataset,'' in
  \emph{2008 IEEE Computer Society Conference on Computer Vision and Pattern
  Recognition Workshops}, June 2008, pp. 1--8.

\bibitem{NIPS2012_4824}
\BIBentryALTinterwordspacing
A.~Krizhevsky, I.~Sutskever, and G.~E. Hinton, ``Imagenet classification with
  deep convolutional neural networks,'' in \emph{Advances in Neural Information
  Processing Systems 25}, F.~Pereira, C.~J.~C. Burges, L.~Bottou, and K.~Q.
  Weinberger, Eds.\hskip 1em plus 0.5em minus 0.4em\relax Curran Associates,
  Inc., 2012, pp. 1097--1105. [Online]. Available:
  \url{http://papers.nips.cc/paper/4824-imagenet-classification-with-deep-convolutional-neural-networks.pdf}
\BIBentrySTDinterwordspacing

\bibitem{298725}
Y.~{Bengio}, P.~{Frasconi}, and P.~{Simard}, ``The problem of learning
  long-term dependencies in recurrent networks,'' in \emph{IEEE International
  Conference on Neural Networks}, March 1993, pp. 1183--1188 vol.3.

\bibitem{Bengio:1994:LLD:2325857.2328340}
\BIBentryALTinterwordspacing
Y.~Bengio, P.~Simard, and P.~Frasconi, ``Learning long-term dependencies with
  gradient descent is difficult,'' \emph{Trans. Neur. Netw.}, vol.~5, no.~2,
  pp. 157--166, Mar. 1994. [Online]. Available:
  \url{http://dx.doi.org/10.1109/72.279181}
\BIBentrySTDinterwordspacing

\bibitem{Hochreiter:1997:LSM:1246443.1246450}
\BIBentryALTinterwordspacing
S.~Hochreiter and J.~Schmidhuber, ``Long short-term memory,'' \emph{Neural
  Comput.}, vol.~9, no.~8, pp. 1735--1780, Nov. 1997. [Online]. Available:
  \url{http://dx.doi.org/10.1162/neco.1997.9.8.1735}
\BIBentrySTDinterwordspacing

\bibitem{DBLP:journals/corr/KingmaB14}
\BIBentryALTinterwordspacing
D.~P. Kingma and J.~Ba, ``Adam: {A} method for stochastic optimization,''
  \emph{CoRR}, vol. abs/1412.6980, 2014. [Online]. Available:
  \url{http://arxiv.org/abs/1412.6980}
\BIBentrySTDinterwordspacing

\bibitem{baseline_feature_based}
T.~M. Emmanuel, ``Using machine learning for real-time activity recognition and
  estimation of energy expenditure,'' 01 2008.

\end{thebibliography}
	
\end{document}